\documentclass{article}

\PassOptionsToPackage{numbers, compress}{natbib}
\usepackage{amsmath}
\usepackage[preprint]{neurips_2022}

\usepackage[utf8]{inputenc} 
\usepackage[T1]{fontenc}    
\usepackage{hyperref}       
\usepackage{url}            
\usepackage{booktabs}       
\usepackage{amsfonts}       
\usepackage{nicefrac}       
\usepackage{microtype}      
\usepackage{xcolor}         
\usepackage[utf8]{inputenc}
\usepackage[english]{babel}
\usepackage[autostyle]{csquotes}
\usepackage{hyperref}
\usepackage{graphicx}
\usepackage{xcolor}         

\title{Edit Everything: A Text-Guided Generative System for Images Editing}

\author{Defeng Xie\thanks{Equal Contributions.}\And
  Ruichen Wang \And
  Jian Ma \AND
  Chen Chen \thanks{Correspondence to: Chen Chen <chenchen4@oppo.com>, Haonan Lu <luhaonan@oppo.com>, Dong Yang <yangdong1@oppo.com or dongyang3-c@my.cityu.edu.hk>.} \And
  Haonan Lu \footnotemark[2] \And
  Dong Yang \footnotemark[2] \AND 
 {\normalfont OPPO Research Institute  }
  \AND
  Fobo Shi \footnotemark[1] \\
  Central China Normal University \\  
  \And
  Xiaodong Lin\\
  Rugster Unversity \\
  }

\begin{document}

\maketitle

\begin{abstract}
We introduce a new generative system called Edit Everything, which can take
image and text inputs and produce image outputs. Edit Everything allows users to
edit images using simple text instructions. Our system designs prompts to guide the
visual module in generating requested images. Experiments demonstrate that Edit
Everything facilitates the implementation of the visual aspects of Stable Diffusion
with the use of Segment Anything model and CLIP. Our system is publicly available
at \textcolor{blue}{\url{https://github.com/DefengXie/Edit_Everything}}.
\end{abstract}

\section{Introduction}

\enquote{While drawing I discover what I really want to say.}

―  Dario Fo

\medskip

Visualization, including images, paintings, shots, illustrations, and photographs, can usually be described with text. However, creating these images often requires specialized skills and a significant time investment \cite{vijayanarasimhan2009s}. Consequently, a generative system has the potential to produce realistic images based on natural language, allowing humans to efficiently generate a wide range of visual content. Additionally, this system offers an unprecedented opportunity for continuous improvement and precise control over image editing, making it a crucial tool for real-world applications. 


Recently, diffusion models have shown promising performances in generating high-quality realistic images \cite{sohl2015deep,rombach2022high,saharia2022photorealistic,ramesh2021zero}. In particular, a text-guided method significantly improves the diversity and fidelity \cite{nichol2021glide}. To address photorealism in the conditional setting, \cite{dhariwal2021diffusion} proposes a classifier to guide diffusion models, allowing them to generate realistic images toward a classifier's label. \cite{ho2022classifier} shows that guidance can be indeed performed by a pure generative model without a classifier, named classifier-free guidance. Classifier-free guidance combines conditional and unconditional score estimates to attain a trade-off between sample quality and diversity similar to that with classifier guidance. Inspired by the ability of guided diffusion models with text, \cite{nichol2021glide} first implements CLIP to guide diffusion models towards text prompt \cite{radford2021learning}. However, compared to the quality of generated images, CLIP guidance cannot compete with classifier-free guidance. Another promising solution is that \cite{zhang2023adding} designs a neural network structure (ControlNet) to control diffusion models and support conditional inputs. Diffusion models can be augmented by ControlNets to generate edge maps, segmentation maps, key points, etc. 

\begin{figure*}[!t]
\centering

\includegraphics[width=14cm]{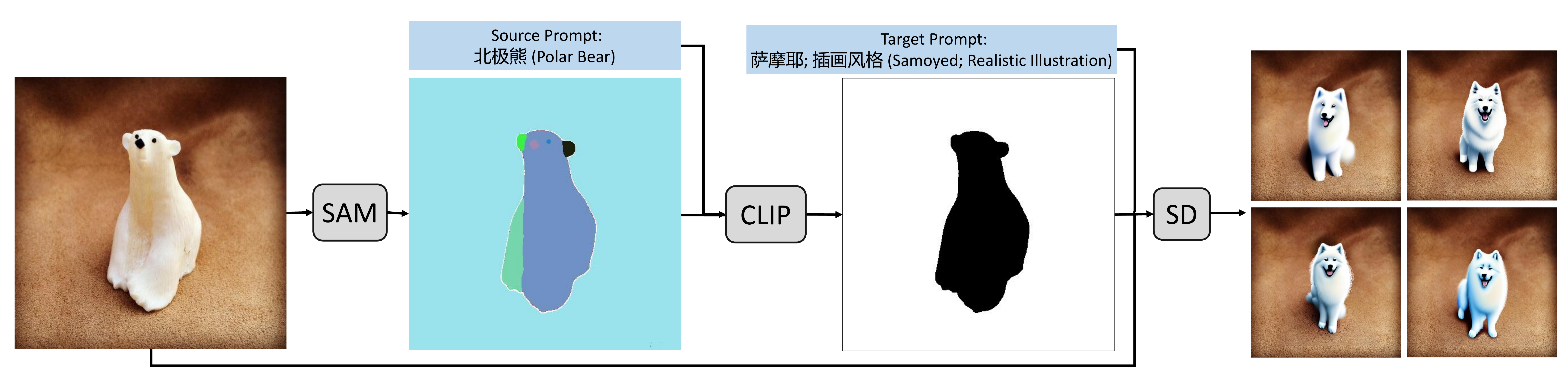}
\caption{The network architecture of Edit Everything. The original image is separated into several segments with the help of Segment Anything model (SAM). Next, These segments are ranked based on the source prompt, and the target segment is chosen based on the highest score calculated by our trained CLIP model. The source prompt is a text that describes the target object and editing styles. Finally, guided by the target prompt, Stable Diffusion (SD) generates the replacement object for the mask segment. This process is seamless and efficient, resulting in high-quality image editing.}
\label{fig1}
\end{figure*}

Inspired by the remarkable performance of ControlNet and CLIP guidance in significantly enhancing the image quality, we leverage Segment Anything model (SAM) and CLIP to guide diffusion models \cite{radford2021learning,kirillov2023segment}. In our work, we create a text-guided generative system called Editing Everything, combining SAM, CLIP and Stable Diffusion (SD) \cite{rombach2022high}. With this framework, we aim to improve the quality of generated images, and provide an efficient and accurate tool for researchers and practitioners in various domains. First, we train a CLIP with 400 million parameters and a SD with 1 billion parameters for Chinese scenarios. Our trained models power Editing Everything, equipping it with the ability to understand Chinese text prompts and guide diffusion models towards text prompts. Second, Editing Everything can readily utilize text prompts for zero-shot generation, while it has difficulty creating realistic images for complex prompts. To address this problem, we propose a solution that this system breaks down complex prompts into smaller entities, which are then replaced in a sequential manner. The system also facilitates iterative sample generation until it matches complex prompts, proving advantageous for users. Third, by leveraging Editing Everything, users can efficiently modify images with different styles and objects. 

\section{Methods}
\subsection{Architecture}

The text-guided generative system, Editing Everything, is composed of three main components: Segment Anything Model (SAM), CLIP, and Stable Diffusion (SD). SAM is used to extract all segments of an image, while CLIP is trained to rank these segments based on a given source prompt. The source prompt describes the interested object. The highest scoring segment is then selected as the target segment. Finally, SD is guided by a target prompt to generate a new object to replace the selected segment. This allows for a precise and personalized method of image editing.

For Chinese scenarios, our CLIP is pre-trained on Chinese corpus and image pairs. And Stable Diffusion is also pre-trained on Chinese corpus. Our generative system has successfully generated realistic images on Chinese scenarios. However, for English scenarios, we implement available open-source CLIP and Stable Diffusion.

\begin{figure*}[!t]
\centering
\includegraphics[width=14cm]{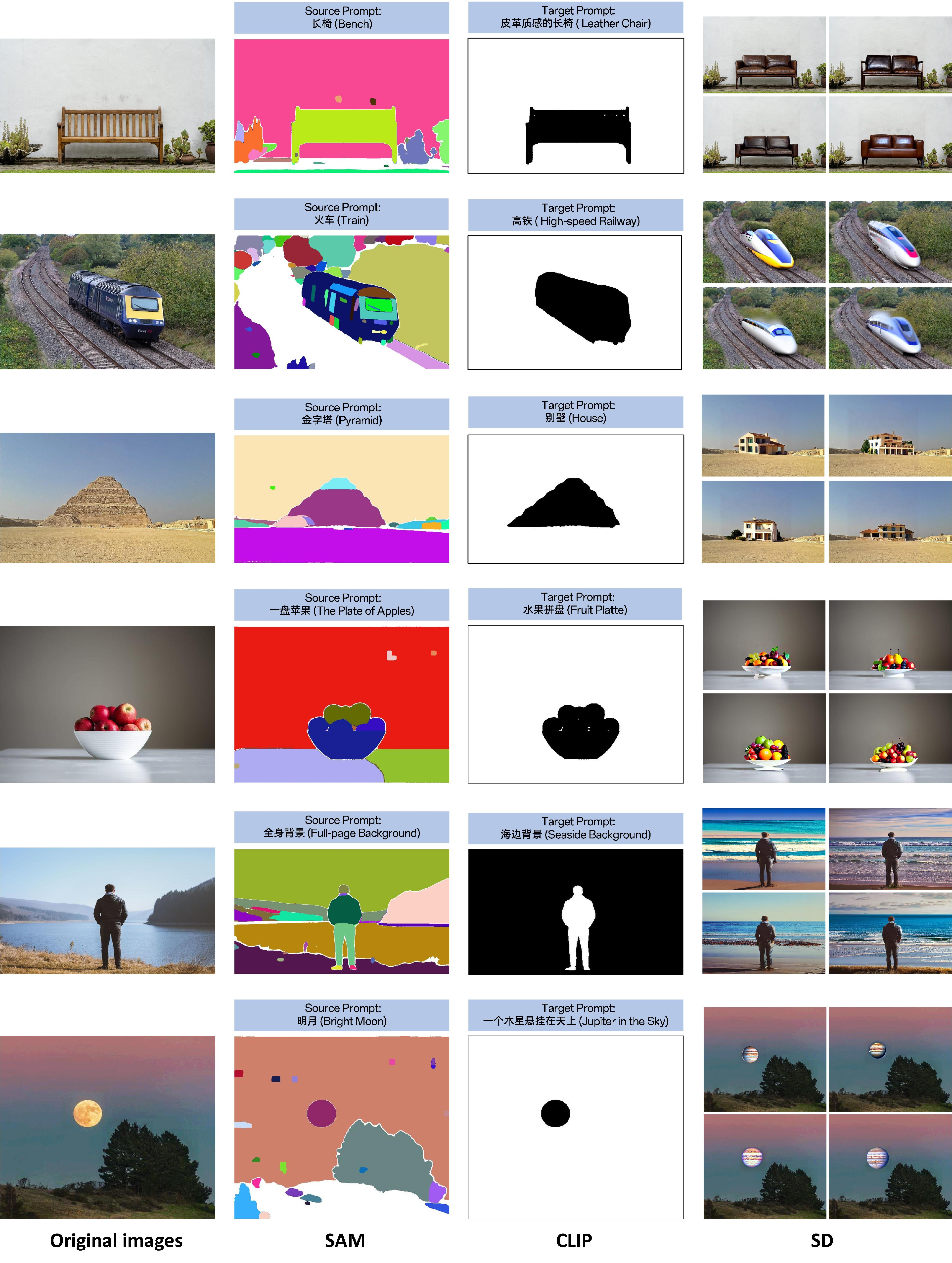}
\caption{Text-guided image editing examples created by Edit Everything. Our advanced system detects the dark region, and erases them by the source target. And then we apply SD to fill it based on the target prompt. Our system is able to produce various styles and seamlessly match the surrounding context.}
\label{fig2}
\end{figure*}

\subsection{Pre-training Data}

Table 1 shows that our pre-training dataset consists of several open sources. Our trained models only use the Chinese corpus and related images. Here are details of these datasets:

\begin{figure*}[!t]
\centering
\includegraphics[width=15cm]{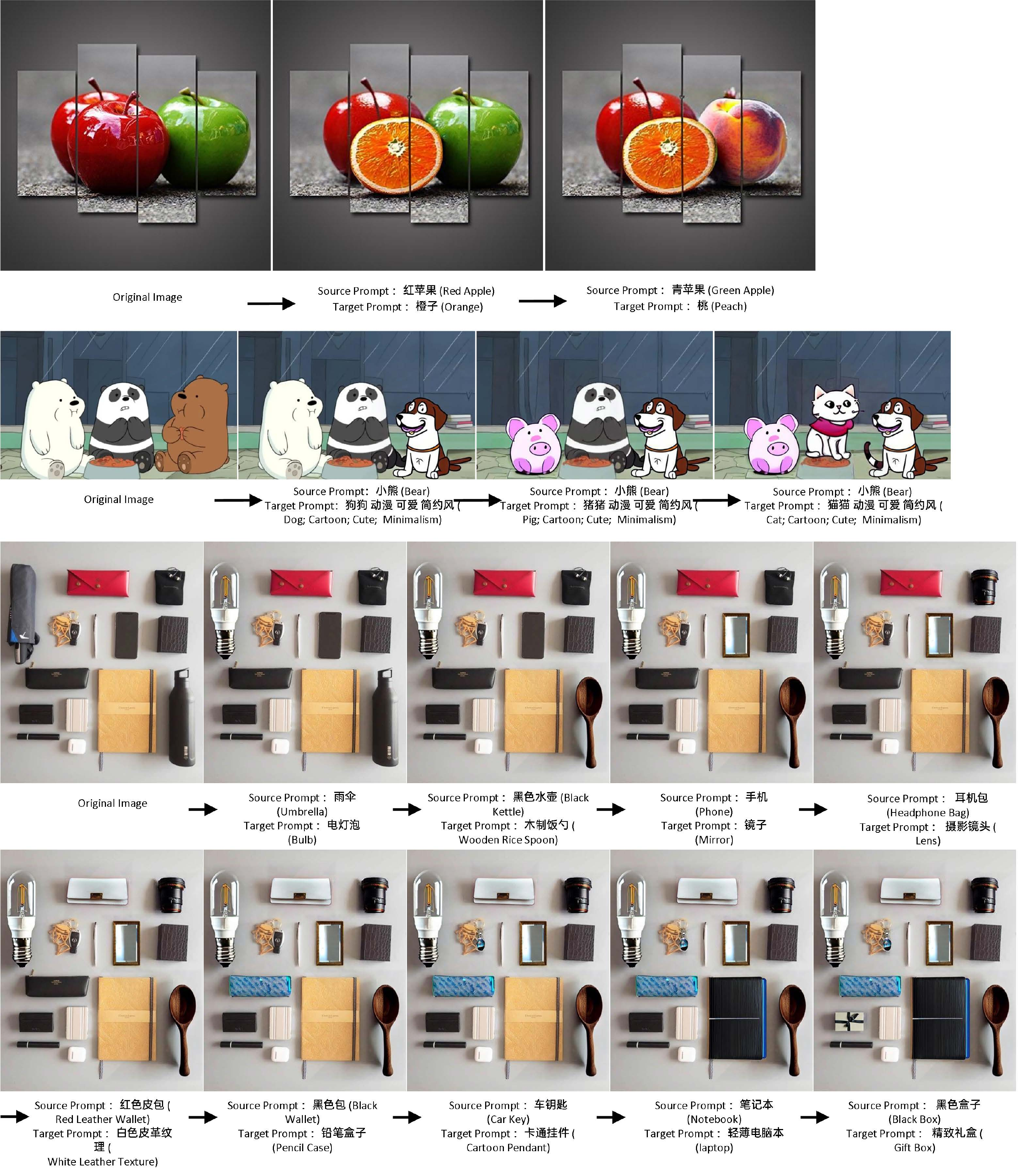}
\caption{Iteratively replacing objects of an image step by step using Editing Everything.}
\label{fig3}
\end{figure*}

\begin{table*}[t]
\vspace{-1.4em}
  \caption{Statistics of pre-training datasets.}
  \centering
  \begin{tabular}{lcc}
    \toprule
    \textbf{Dataset} &\textbf{Proportion (\%)} & \textbf{Number of image-text pairs (million)}\\
    \hline
    Wukong & 33.0& 100 \\
    Zero and R2D2 & 7.6& 23 \\
    Laion-5b & 26.4& 80 \\
    Crawled Data & 33.0& 100 \\
    \bottomrule
  \end{tabular}
  \label{table1}
\end{table*}

\textbf{Wukong.} Wukong is the largest Chinese cross-modal dataset, an enormous collection of 100 million Chinese image-text pairs from the web \cite{gu2022wukong}. This dataset is designed for large-scale experiments and serves as a valuable resource for researchers in language-vision fields. In our training, we use 100 million image-text pairs. 

\textbf{Zero and R2D2.} Zero and R2D2 is a large-scale cross-modal dataset, which contains the public dataset ZERO-Corpus and five human-annotated datasets designed for downstream tasks \cite{xie2022zero}. It has 250 million images paired with 750 million text. Based on Zero and R2D2, we can extract 23 million high-quality Chinese image-text pairs for our training.

\textbf{Laion-5b.} Laion-5b is a large-scale multi-modal dataset consisting of 5.85 billion CLIP-filtered image-text pairs, of which 2.32B are English language \cite{schuhmann2022laion}. We have processed 80 million Chinese image-text pairs by filtering and pre-processing the data. 

\textbf{Crawled Data.} We have crawled an extensive collection of 100 million image-text pairs from the web. This dataset covers various domains, such as shopping, common knowledge, tools, etc. However, we regret to inform that due to commercial restrictions, we cannot release this data to the public.

\begin{figure*}[!t]
\centering
\includegraphics[width=14cm]{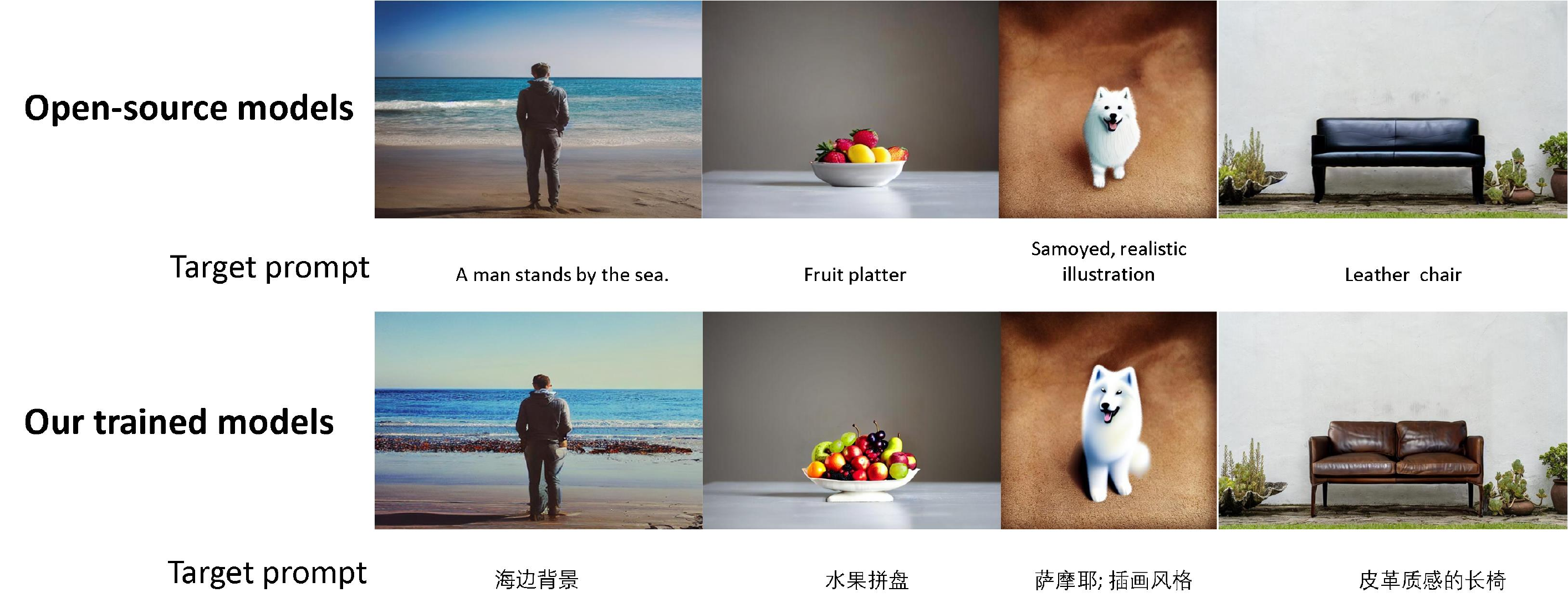}
\caption{Comparisons of images generated by open-source models and our trained models. Our models could support Chinese inputs.}
\label{fig4}
\end{figure*}

\subsection{Implementation}

CLIP is composed of a text encode and an image encode. The text encode is a Chinese BERT with the WordPiece \cite{devlin2018bert}. The image encode utilizes a ViT-L/14 trained by OpenAI's CLIP \cite{dosovitskiy2020image,radford2021learning}. Stable Diffusion is a powerful model that combines VAE \cite{kingma2013auto,razavi2019generating}, UNet \cite{dhariwal2021diffusion}, and text encoder to achieve exceptional performance. The VAE parameters are kept frozen during the training process. The UNet is designed to learn a data distribution, $p(x)$, by gradually denoising a normally distributed variable, allowing it to learn the reverse process of a fixed Markov Chain of length $T$.

Our trained models use the AdamW optimizer \cite{loshchilov2017decoupled}, with the following hyper parameters: $\beta_1=0.9,\beta_2 = 0.99$. To optimize our training process, we implement a linear learning rate schedule, and apply a weight decay of 0.1 and gradient clipping at 1.0. In addition, warm-up steps are set to 2000 and the batch size is 180. To reduce computation costs, we run them on 40 Tesla V100s.

\section{Main Results}

\subsection{Simple Prompts}

In Figure~\ref{fig2}, we observe that Editing Everything with text guidance is capable of editing any object within an image and generating a diverse range of realistic images. It can also seamlessly match different styles of illustrations, such as realistic or painted styles. In editing tasks, Editing Everything can effectively modify existing images using source prompts, and insert new objects, shadows, and reflections based on target prompts. The results are highly realistic.

\subsection{Complicated Prompts}

In Figure~\ref{fig3}. we demonstrate the iterative creation of images using Editing Everything based on complex prompts. This process involves the step-by-step replacement of source objects with target objects. While this pipeline may not be efficient, it produces highly accurate control.

\subsection{Further Comparisons}

To further demonstrate the performance of our system, we present a comparison between our trained models and open-source models in Figure~\ref{fig4}. Our trained models consist of open-source SAM, trained CLIP, and trained Stable Diffusion. In contrast, open-source models contain open-source SAM, Taiyi-CLIP \cite{Fengshenbang-LM}, and trained stable diffusion \cite{rombach2022high}. Notably, our trained models are capable of accepting Chinese text inputs. Moreover, due to crawled high-quality image-text datasets, our models outperform open-source models.

\section{Limitations}

Our generative system, Editing Everything, consists of SAM, CLIP, and SD models. Their architecture is not modified in any way. To enhance their performance in Chinese scenarios, we trained these models on our own crawled dataset.

\section{Conclusion}

In this paper, we propose a new generative system, namely Editing Everything, to design text prompts to help Stable Diffusion to generate images toward target prompts. This generative system implements Segment Anything Model, CLIP, Stable Diffusion. Based on our trained models, this work firstly provides an efficient solution for Chinese scenarios. Compared to open-source models, our generative system achieve a great performance for image editing.

\section{Acknowledgements}

We would like to acknowledge the contribution of Miss. Wenjing Feng, who crawled the high-quality image-text pairs used in this work.

\bibliography{nips2022}

\begin{thebibliography}{20}
\providecommand{\natexlab}[1]{#1}
\providecommand{\url}[1]{\texttt{#1}}
\expandafter\ifx\csname urlstyle\endcsname\relax
  \providecommand{\doi}[1]{doi: #1}\else
  \providecommand{\doi}{doi: \begingroup \urlstyle{rm}\Url}\fi

\bibitem[Devlin et~al.(2018)Devlin, Chang, Lee, and Toutanova]{devlin2018bert}
Jacob Devlin, Ming-Wei Chang, Kenton Lee, and Kristina Toutanova.
\newblock Bert: Pre-training of deep bidirectional transformers for language
  understanding.
\newblock \emph{arXiv preprint arXiv:1810.04805}, 2018.

\bibitem[Dhariwal and Nichol(2021)]{dhariwal2021diffusion}
Prafulla Dhariwal and Alexander Nichol.
\newblock Diffusion models beat gans on image synthesis.
\newblock \emph{Advances in Neural Information Processing Systems},
  34:\penalty0 8780--8794, 2021.

\bibitem[Dosovitskiy et~al.(2020)Dosovitskiy, Beyer, Kolesnikov, Weissenborn,
  Zhai, Unterthiner, Dehghani, Minderer, Heigold, Gelly,
  et~al.]{dosovitskiy2020image}
Alexey Dosovitskiy, Lucas Beyer, Alexander Kolesnikov, Dirk Weissenborn,
  Xiaohua Zhai, Thomas Unterthiner, Mostafa Dehghani, Matthias Minderer, Georg
  Heigold, Sylvain Gelly, et~al.
\newblock An image is worth 16x16 words: Transformers for image recognition at
  scale.
\newblock \emph{arXiv preprint arXiv:2010.11929}, 2020.

\bibitem[Gu et~al.(2022)Gu, Meng, Lu, Hou, Minzhe, Liang, Yao, Huang, Zhang,
  Jiang, et~al.]{gu2022wukong}
Jiaxi Gu, Xiaojun Meng, Guansong Lu, Lu~Hou, Niu Minzhe, Xiaodan Liang, Lewei
  Yao, Runhui Huang, Wei Zhang, Xin Jiang, et~al.
\newblock Wukong: A 100 million large-scale chinese cross-modal pre-training
  benchmark.
\newblock \emph{Advances in Neural Information Processing Systems},
  35:\penalty0 26418--26431, 2022.

\bibitem[Ho and Salimans(2022)]{ho2022classifier}
Jonathan Ho and Tim Salimans.
\newblock Classifier-free diffusion guidance.
\newblock \emph{arXiv preprint arXiv:2207.12598}, 2022.

\bibitem[IDEA-CCNL(2021)]{Fengshenbang-LM}
IDEA-CCNL.
\newblock Fengshenbang-lm.
\newblock \url{https://github.com/IDEA-CCNL/Fengshenbang-LM}, 2021.

\bibitem[Kingma and Welling(2013)]{kingma2013auto}
Diederik~P Kingma and Max Welling.
\newblock Auto-encoding variational bayes.
\newblock \emph{arXiv preprint arXiv:1312.6114}, 2013.

\bibitem[Kirillov et~al.(2023)Kirillov, Mintun, Ravi, Mao, Rolland, Gustafson,
  Xiao, Whitehead, Berg, Lo, et~al.]{kirillov2023segment}
Alexander Kirillov, Eric Mintun, Nikhila Ravi, Hanzi Mao, Chloe Rolland, Laura
  Gustafson, Tete Xiao, Spencer Whitehead, Alexander~C Berg, Wan-Yen Lo, et~al.
\newblock Segment anything.
\newblock \emph{arXiv preprint arXiv:2304.02643}, 2023.

\bibitem[Loshchilov and Hutter(2017)]{loshchilov2017decoupled}
Ilya Loshchilov and Frank Hutter.
\newblock Decoupled weight decay regularization.
\newblock \emph{arXiv preprint arXiv:1711.05101}, 2017.

\bibitem[Nichol et~al.(2021)Nichol, Dhariwal, Ramesh, Shyam, Mishkin, McGrew,
  Sutskever, and Chen]{nichol2021glide}
Alex Nichol, Prafulla Dhariwal, Aditya Ramesh, Pranav Shyam, Pamela Mishkin,
  Bob McGrew, Ilya Sutskever, and Mark Chen.
\newblock Glide: Towards photorealistic image generation and editing with
  text-guided diffusion models.
\newblock \emph{arXiv preprint arXiv:2112.10741}, 2021.

\bibitem[Radford et~al.(2021)Radford, Kim, Hallacy, Ramesh, Goh, Agarwal,
  Sastry, Askell, Mishkin, Clark, et~al.]{radford2021learning}
Alec Radford, Jong~Wook Kim, Chris Hallacy, Aditya Ramesh, Gabriel Goh,
  Sandhini Agarwal, Girish Sastry, Amanda Askell, Pamela Mishkin, Jack Clark,
  et~al.
\newblock Learning transferable visual models from natural language
  supervision.
\newblock In \emph{International conference on machine learning}, pages
  8748--8763. PMLR, 2021.

\bibitem[Ramesh et~al.(2021)Ramesh, Pavlov, Goh, Gray, Voss, Radford, Chen, and
  Sutskever]{ramesh2021zero}
Aditya Ramesh, Mikhail Pavlov, Gabriel Goh, Scott Gray, Chelsea Voss, Alec
  Radford, Mark Chen, and Ilya Sutskever.
\newblock Zero-shot text-to-image generation.
\newblock In \emph{International Conference on Machine Learning}, pages
  8821--8831. PMLR, 2021.

\bibitem[Razavi et~al.(2019)Razavi, Van~den Oord, and
  Vinyals]{razavi2019generating}
Ali Razavi, Aaron Van~den Oord, and Oriol Vinyals.
\newblock Generating diverse high-fidelity images with vq-vae-2.
\newblock \emph{Advances in neural information processing systems}, 32, 2019.

\bibitem[Rombach et~al.(2022)Rombach, Blattmann, Lorenz, Esser, and
  Ommer]{rombach2022high}
Robin Rombach, Andreas Blattmann, Dominik Lorenz, Patrick Esser, and Bj{\"o}rn
  Ommer.
\newblock High-resolution image synthesis with latent diffusion models.
\newblock In \emph{Proceedings of the IEEE/CVF Conference on Computer Vision
  and Pattern Recognition}, pages 10684--10695, 2022.

\bibitem[Saharia et~al.(2022)Saharia, Chan, Saxena, Li, Whang, Denton,
  Ghasemipour, Gontijo~Lopes, Karagol~Ayan, Salimans,
  et~al.]{saharia2022photorealistic}
Chitwan Saharia, William Chan, Saurabh Saxena, Lala Li, Jay Whang, Emily~L
  Denton, Kamyar Ghasemipour, Raphael Gontijo~Lopes, Burcu Karagol~Ayan, Tim
  Salimans, et~al.
\newblock Photorealistic text-to-image diffusion models with deep language
  understanding.
\newblock \emph{Advances in Neural Information Processing Systems},
  35:\penalty0 36479--36494, 2022.

\bibitem[Schuhmann et~al.(2022)Schuhmann, Beaumont, Vencu, Gordon, Wightman,
  Cherti, Coombes, Katta, Mullis, Wortsman, et~al.]{schuhmann2022laion}
Christoph Schuhmann, Romain Beaumont, Richard Vencu, Cade Gordon, Ross
  Wightman, Mehdi Cherti, Theo Coombes, Aarush Katta, Clayton Mullis, Mitchell
  Wortsman, et~al.
\newblock Laion-5b: An open large-scale dataset for training next generation
  image-text models.
\newblock \emph{arXiv preprint arXiv:2210.08402}, 2022.

\bibitem[Sohl-Dickstein et~al.(2015)Sohl-Dickstein, Weiss, Maheswaranathan, and
  Ganguli]{sohl2015deep}
Jascha Sohl-Dickstein, Eric Weiss, Niru Maheswaranathan, and Surya Ganguli.
\newblock Deep unsupervised learning using nonequilibrium thermodynamics.
\newblock In \emph{International Conference on Machine Learning}, pages
  2256--2265. PMLR, 2015.

\bibitem[Vijayanarasimhan and Grauman(2009)]{vijayanarasimhan2009s}
Sudheendra Vijayanarasimhan and Kristen Grauman.
\newblock What's it going to cost you?: Predicting effort vs. informativeness
  for multi-label image annotations.
\newblock In \emph{2009 IEEE conference on computer vision and pattern
  recognition}, pages 2262--2269. IEEE, 2009.

\bibitem[Xie et~al.(2022)Xie, Cai, Song, Li, Kong, Wu, Morimitsu, Yao, Wang,
  Leng, et~al.]{xie2022zero}
Chunyu Xie, Heng Cai, Jianfei Song, Jincheng Li, Fanjing Kong, Xiaoyu Wu,
  Henrique Morimitsu, Lin Yao, Dexin Wang, Dawei Leng, et~al.
\newblock Zero and r2d2: A large-scale chinese cross-modal benchmark and a
  vision-language framework.
\newblock \emph{arXiv preprint arXiv:2205.03860}, 2022.

\bibitem[Zhang and Agrawala(2023)]{zhang2023adding}
Lvmin Zhang and Maneesh Agrawala.
\newblock Adding conditional control to text-to-image diffusion models.
\newblock \emph{arXiv preprint arXiv:2302.05543}, 2023.

\end{thebibliography}


\appendix

\end{document}